\documentclass[sigconf,screen]{acmart}
\AtBeginDocument{%
  }

\setcopyright{acmlicensed}
\copyrightyear{2025}
\acmYear{2025}
\acmDOI{XXXXXXX.XXXXXXX}
\acmConference[ACM SIGSPATIAL '25]{33rd ACM SIGSPATIAL International Conference on Advances in Geographic Information Systems}{November 3--6, 2025}{Minneapolis, MN, USA}

\renewcommand\footnotetextcopyrightpermission[1]{} 

\begin{document}


\title[TerraMAE]{TerraMAE: Learning Spatial-Spectral Representations from Hyperspectral Earth Observation Data via Adaptive Masked Autoencoders}


\author{Tanjim Bin Faruk}
\email{tanjim@colostate.edu}
\affiliation{%
  \institution{Colorado State University}
  \city{Fort Collins}
  \state{Colorado}
  \country{USA}
}

\author{Abdul Matin}
\affiliation{%
  \institution{Colorado State University}
  \city{Fort Collins}
  \state{Colorado}
  \country{USA}
}
\email{amatin@colostate.edu}

\author{Shrideep Pallickara}
\affiliation{%
  \institution{Colorado State University}
  \city{Fort Collins}
  \state{Colorado}
  \country{USA}
}
\email{shrideep@colostate.edu}

\author{Sangmi Lee Pallickara}
\affiliation{%
  \institution{Colorado State University}
  \city{Fort Collins}
  \state{Colorado}
  \country{USA}
}
\email{sangmi@colostate.edu}

\renewcommand{\shortauthors}{Faruk et al.}

\begin{abstract}
Hyperspectral satellite imagery offers sub-30 m views of Earth in hundreds of contiguous spectral bands, enabling fine-grained mapping of soils, crops, and land cover. While self-supervised Masked Autoencoders excel on RGB and low-band multispectral data, they struggle to exploit the intricate spatial–spectral correlations in 200\texttt{+} band hyperspectral images. We introduce TerraMAE, a novel HSI encoding framework specifically designed to learn highly representative spatial-spectral embeddings for diverse geospatial analyses. TerraMAE features an adaptive channel grouping strategy, based on statistical reflectance properties to capture spectral similarities, and an enhanced reconstruction loss function that incorporates spatial and spectral quality metrics. We demonstrate TerraMAE's effectiveness through superior spatial-spectral information preservation in high-fidelity image reconstruction. Furthermore, we validate its practical utility and the quality of its learned representations through strong performance on three key downstream geospatial tasks: crop identification, land cover classification, and soil texture prediction.
\end{abstract}





\maketitle

\begin{center}
\textit{© Author 2025. This is the author's version of the work. It is posted here for your personal use. Not for redistribution. The definitive Version of Record will be published in Proceedings of the 33rd ACM SIGSPATIAL International Conference on Advances in Geographic Information Systems, ACM, 2025.}
\end{center}

\section{Introduction}
\label{sec:intro}

Geographic processes such as soil formation, vegetation growth, and land use change create complex spatial-spectral signatures across Earth's surface. These signatures underpin critical analyses in agriculture, climate adaptation, and environmental monitoring. However, extracting these signatures at scale remains challenging due to the intricate interplay between spatial variation—how features change across geographic space—and spectral variation—how surface reflectance varies across the electromagnetic spectrum.

Hyperspectral Satellite Images (HSIs) offer unprecedented capability to analyze these geographic signatures by capturing reflectance across hundreds of narrow, contiguous spectral bands, often exceeding 200 channels. This stands in contrast to the more limited range of multispectral images, which typically include between 3 and 11 wider bands \cite{hagen-kudenov}. The rich spectral information in HSIs supports fine-grained analyses such as identifying plant and tree species, monitoring air quality, and assessing vegetation health. Prominent hyperspectral satellite missions include EnMAP (used in this study) \cite{STORCH2023113632}, PRISMA \cite{COGLIATI2021112499}, and Planet's Tanager-1 \cite{PBC2024}.

The challenge lies in developing computational methods that can effectively model the dual spatial-spectral nature of geographic phenomena in HSI data. Traditional remote sensing methods often process spatial and spectral dimensions independently \cite{10168240, rs16010193}, potentially overlooking critical interactions. For instance, analyzing spectral bands in isolation may miss how reflectance patterns change across space due to variations in land cover types or soil properties. Supervised deep learning methods have shown promise \cite{Rolf2021, su151310543}, but they are severely limited by the scarcity of labeled geographic data---particularly for HSI, where field collection is expensive and logistically challenging across large geographic extents.


Self-supervised learning (SSL) offers a way forward by learning representations from the vast archives of unlabeled satellite imagery and transferring them to specific downstream tasks \cite{9462394, technologies9010002}. However, the assumptions built into existing self-supervised methods, primarily designed for RGB imagery, do not always carry over to satellite imagery. In satellite imagery, each pixel encodes both spatial and spectral information, and neighboring pixels often exhibit structured patterns tied to meaningful physical characteristics. The challenge becomes even more complex with HSI, where the number of spectral channels far exceeds the three bands found in standard RGB images, introducing complex high-dimensional dependencies that conventional RGB-focused methods are ill-suited to capture.




Recent deep learning frameworks in remote sensing have largely focused on multispectral data \cite{10377166, SATMAE, Li_2024_CVPR}. Some newer models have extended to HSI, though their focus has often remained on classification tasks \cite{10678397, 10607879}. While classification performance is an important benchmark, it captures only one aspect of geospatial analysis. In contrast, masked autoencoder (MAE) provides a self-supervised framework that learns general-purpose spatial-spectral representations through a reconstruction objective, which can then be transferred to a range of downstream geospatial tasks that require a finer-grained understanding of the spectral and spatial variations. 

MAEs offer a compelling solution to one of the central challenges in hyperspectral remote sensing: extracting meaningful spatial-spectral representations from vast, unlabeled datasets.  Traditional supervised approaches require extensive labeled data, which proves untenable for HSI due to the high cost and logistical difficulty of field annotation across large geographic extents. The complexity of HSI—characterized by hundreds of spectral channels—further amplifies the challenge, as it marks transitions in soil composition, vegetation health, or anthropogenic perturbations—patterns that often evade discrete classification but carry important signals about underlying environmental processes.

MAEs respond directly to this challenge by reconstructing masked portions of the input from surrounding context, requiring the model to capture complex spatial and spectral dependencies. This process generates features that are not only informative for a wide range of downstream tasks but also generalize robustly across different landscapes, sensors, and seasons. By enabling effective training on large-scale, unlabeled satellite data, MAEs unlock the potential of geospatial analysis to match the complexity of Earth surface phenomena.

\subsection{Research Questions}
The overarching research question that we explore in this study is: \textit{How can self-supervised MAEs be adapted to effectively pretrain models on HSI, given its complex spatial-spectral structure and the limited availability of labeled data?} We break this down into two focused research questions:

\textbf{RQ-1}: \textit{How can reconstruction objectives in MAEs be designed to jointly preserve spatial structure and spectral fidelity when applied to HSI?} This concerns the design of the loss function at the heart of pretraining—one that must account for the dual structure of the data.

\textbf{RQ-2}: \textit{Given the high number of correlated spectral channels in HSI (200+), what strategies for channel grouping enable MAEs to scale effectively to HSI, and in what ways can adaptive, data-driven groupings outperform fixed, wavelength-based partitions?} The focus here is on organizing spectral channels in a way that reflects their inherent statistical structure, rather than imposing rigid external boundaries.

\subsection{Approach Summary}
In this study, we introduce TerraMAE, a MAE framework tailored specifically for HSI. TerraMAE enhances pretraining quality by incorporating a reconstruction loss that jointly preserves spatial structure and spectral fidelity—two defining characteristics of HSI. By modeling these dimensions together during pretraining, the encoder learns robust spatial-spectral representations, enabling effective reconstruction and transfer to downstream geospatial tasks, even in the absence of labeled data.

A core innovation of TerraMAE is its adaptive channel grouping mechanism, which addresses the limitations of static, wavelength-based spectral partitions. Previous work—such as SatMAE \cite{SATMAE}—has shown that grouping channels can assist in pretraining for multispectral imagery, which typically includes far fewer bands ($\sim$13 or so). While fixed groupings may suffice for such lower-dimensional multispectral imagery, they do not scale effectively to HSI, where the number of channels often exceeds 200. Static partitions risk overlooking meaningful spectral relationships, particularly among non-adjacent or spectrally distant channels, and lack the flexibility needed to accommodate variability across datasets. TerraMAE circumvents these constraints by clustering channels based on the empirical distribution of reflectance values within the pretraining dataset. This data-driven approach more effectively captures latent spectral relationships and complex spatial-spectral interactions, leading to more faithful reconstructions and stronger, more transferable representations for downstream tasks.


\subsection{Paper Contributions}

This work makes the following key contributions:

\begin{itemize}
    \item We propose TerraMAE, a pretraining framework based on masked autoencoders that leverages an adaptive channel grouping strategy based on spectral similarity. This enables the model to capture meaningful relationships both within and across coherent sets of hyperspectral channels.
    
    \item We introduce a novel reconstruction loss tailored to the spatial-spectral complexity of high-dimensional HSI. By jointly optimizing for spatial structure (via Structural Similarity Index) and spectral fidelity (via Spectral Information Divergence), our loss function guides the model toward more informative and transferable representations.
    
    \item We conduct comprehensive empirical evaluations on diverse geospatial prediction tasks—including soil texture regression, crop type classification, and land cover mapping—demonstrating that TerraMAE consistently outperforms a strong baseline MAE in both reconstruction accuracy and downstream task performance.
\end{itemize}

\textbf{Translational Impacts}: By enabling more accurate pretraining on unlabeled hyperspectral data, TerraMAE facilitates reliable extraction of information about vegetation health, land cover, and atmospheric conditions in regions where labeled data are limited or unavailable. We posit that TerraMAE's data-driven design makes it broadly applicable across satellite platforms and geographic regions. This generalizability is suited for domains such as precision agriculture, biodiversity assessment, and sustainable urban planning—where high spectral resolution is essential but labeled supervision remains challenging to obtain at scale.


\begin{figure*}[!htbp]
  \centering
    \includegraphics[height=8cm, width=0.8\textwidth]{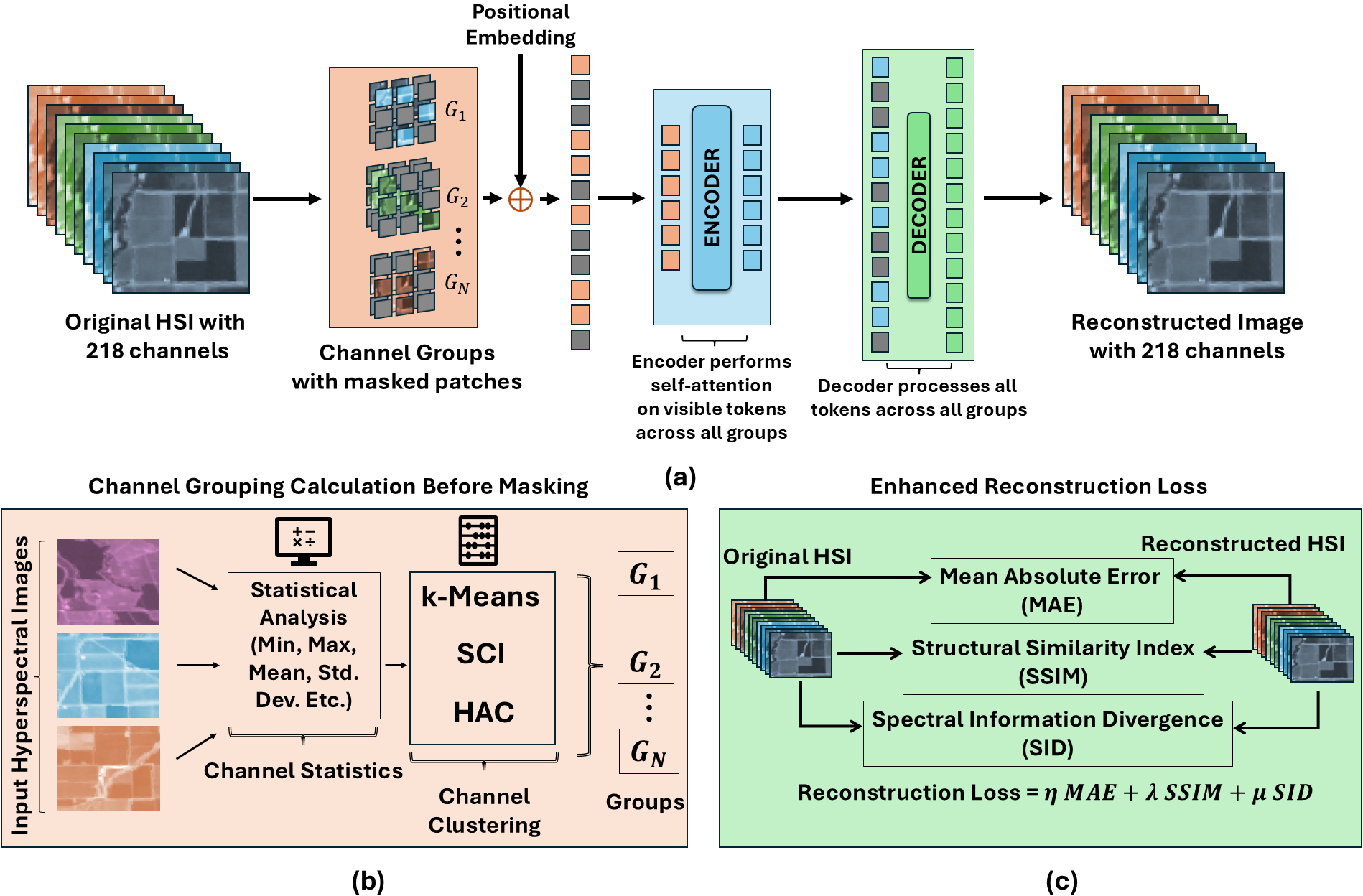}
  \caption{By incorporating adaptive channel grouping (b) and an enriched loss function (c) for 218-band hyperspectral imagery, TerraMAE (a) produces superior latent representations well-suited for diverse remote sensing tasks.}
  \Description{TerraMAE Architecture}
  \label{fig:architecture}
\end{figure*}

\section{Related Work}
\label{sec:rel-work}

The application of SSL and foundation models to remote sensing imagery, particularly high-dimensional HSI, has gained significant traction in recent years. This section reviews advances in MAE-based SSL for satellite data, emerging HSI foundation models, and broader deep learning strategies for HSI feature extraction. We contextualize TerraMAE within this landscape, emphasizing its methodological innovations for spectral grouping and loss design.

\subsection{Masked Autoencoders for Remote Sensing}
The MAE framework \cite{9879206} has demonstrated strong representation learning capabilities in vision tasks and has been adapted for remote sensing. Early adaptations such as SatMAE \cite{SATMAE} introduced temporal and wavelength-based masking strategies for Sentinel-2 imagery, paving the way for SSL in Earth observation. SatMAE++ \cite{satmae++} extended this with scalable pretraining for downstream classification tasks, while ScaleMAE \cite{Reed_2023_ICCV} focused on resolution-aware representation learning. These efforts primarily target multispectral data.

For HSI, which often includes over 200 spectrally correlated bands, specialized adaptations are required. SpectralMAE \cite{s23073728} emphasized spectral reconstruction using fixed band-wise masking, while SSMAE \cite{10314566} combined transformer-based global encoding with CNN-based local feature extraction. However, these methods have not been extensively evaluated on geospatial tasks and do not fully leverage the semantic structure inherent in hyperspectral data. TerraMAE addresses these limitations through specialized masking and loss designs tailored for HSI characteristics.

\subsection{Foundation Models for HSI} Recent work has explored large-scale foundation models to provide generalizable representations for HSI analysis. SpectralEarth \cite{braham2024spectralearthtraininghyperspectralfoundation}, trained on $\sim$538k EnMAP patches using multiple SSL paradigms (MAE, MoCo-V2, DINO), shows strong performance on crop segmentation and land cover mapping. However, the lack of publicly available weights and task standardization limits reproducibility and broader adoption.

DOFA \cite{xiong2024neuralplasticityinspiredmultimodalfoundation} adopts a multimodal strategy, incorporating over 11,000 HSI samples from HySpecNet-11k \cite{10283385} for tasks like wildfire detection and soil parameter estimation. Yet, the HSI branch remains under development, and the dataset lacks geolocation metadata, posing challenges for spatially explicit applications such as precision agriculture and land use modeling.

HyperSIGMA \cite{10949864} leverages the HyperGlobal-450K corpus, comprising EO-1 Hyperion and Gaofen-5B imagery. Although pretrained ViT-L weights are released, the proprietary nature of the dataset and spectral differences across sensors complicate reproducibility and transferability to EnMAP-based applications.

In contrast, TerraMAE demonstrates that targeted architectural design, rather than data scale alone, can yield strong hyperspectral representations. Trained on $\sim$22,000 publicly available EnMAP tiles, it incorporates spectral clustering-informed masking and a spatially guided spectral reconstruction loss. These innovations enable effective transfer to key geospatial tasks such as soil texture prediction, crop type classification, and land cover classification.

\subsection{Deep Learning Approaches for HSI Feature Extraction} Before the rise of foundation models, deep learning approaches for HSI focused on spectral-spatial feature learning via custom architectures. 3D-ADNet \cite{shi2021hyperspectral} used attention mechanisms for denoising and regression, while HSI-DeNet \cite{he2021convolutional} applied convolutional autoencoders for restoration and classification. Though effective in task-specific settings, these supervised methods required extensive labeled data and lacked generalization across domains.

TerraMAE builds on this legacy by adopting a self-supervised paradigm that removes the need for manual annotation and directly aligns model design with the spectral and spatial properties of HSI. This shift reflects a broader evolution from task-bound networks to pretraining-driven frameworks that support scalable, multi-task geospatial learning.

\section{Methodology}

The overall structure of TerraMAE is presented in Figure \ref{fig:architecture}. Although it builds upon the standard MAE framework, its design responds directly to the unique challenges of HSI: high spectral dimensionality, complex spatial-spectral relationships, and limited access to labeled data. To address these, TerraMAE introduces two principal innovations. The first is a data-driven channel grouping strategy, in which spectral bands are clustered according to the statistical properties of their reflectance values. This approach enhances the model's ability to learn meaningful relationships across both adjacent and distant spectral channels (Section \ref{subsec:channel_grouping}). The second is a composite loss function that integrates Mean Absolute Error, Structural Similarity Index, and Spectral Information Divergence—each contributing to the model's capacity to preserve both structural coherence and spectral fidelity during the reconstruction process (Section \ref{subsec:loss_function}). 

\subsection{Adaptive Channel Grouping with Independent Masking - [RQ-2]}
\label{subsec:channel_grouping}

The conventional practice in MAEs is to apply a uniform mask across all channels of an input image. While this approach has proven effective for natural image datasets (such as ImageNet, which consists primarily of RGB images), it becomes far less suitable in the context of HSI. Unlike RGB images, HSIs often comprise more than 200 spectral bands, a scale at which uniform masking begins to obscure more than it reveals. The central issue lies in the increased likelihood that spatially informative regions in a satellite image will be entirely masked across all channels, thereby depriving the model of critical spectral and spatial cues. This problem is especially acute in HSIs, where subtle spectral signatures and fine-grained spatial features carry essential information.

To overcome this limitation, TerraMAE introduces an adaptive channel grouping strategy in combination with independent masking within each group. Rather than masking the entire spectral stack uniformly, the model applies masks independently within groups of spectrally similar channels—while still maintaining a consistent overall masking ratio across channels and images. This strategy serves two purposes. First, it allows the model to attend more effectively to meaningful spectral features within each group, thereby enhancing its capacity to learn fine-grained spatial-spectral relationships. Second, it reduces the computational complexity of learning across more than 200 channels, avoiding the inefficiencies and risks of overfitting associated with full-spectrum modeling.

At the heart of this grouping strategy lies the Spectral Comparison Index (SCI)—a similarity metric that we introduce to quantify the relationship between HSI channels by comparing their reflectance values across corresponding spatial regions. Specifically, for two spectral bands $i$ and $j$, the SCI is calculated as:

\begin{equation}
    SCI_{i, j} = 1 - \frac{|I_i - I_j|}{I_i + I_j + \epsilon} 
    \label{eq:sci}
\end{equation}

where $I_i$ and $I_j$ are the reflectance values of bands $i$ and $j$ respectively, and $\epsilon$ is a small constant to prevent division by zero. The SCI values range from $0$ to $1$, with higher values indicating greater similarity between the bands.

To compute SCI between channels, we begin by computing a mean reflectance image for each HSI channel. This is achieved by averaging the normalized reflectance values at each spatial location across the entire training dataset. The result is a set of per-channel mean images, each capturing the typical spatial distribution of reflectance for its corresponding band. These images serve as representative profiles for comparing spectral behavior across space.

Next, we evaluate the SCI between each pair of channels by performing an element-wise comparison of their mean reflectance images. Specifically, for channels $i$ and $j$, we apply the SCI formula (as defined in Equation~\ref{eq:sci}) at each spatial location, generating a two-dimensional similarity map. By taking into account the spatial distribution of reflectance, this map reflects the per-pixel spectral similarity of the two channels across the spatial domain. 

To summarize this map into a single inter-channel similarity score, we calculate $SCI_{prod}$ as the product of the mean SCI ($SCI_{\mu}$) across the spatial map and a stability factor ($1 - SCI_{\sigma}$):

\begin{equation}
    SCI_{prod} = SCI_{\mu} \times (1 - SCI_{\sigma}) 
    \label{eq:sci_prod}
\end{equation}

where $SCI_{\sigma}$ is the standard deviation of the SCI values from the 2D similarity map. The stability term ($1 - SCI_{\sigma}$) acts as a differential weighting mechanism; band pairs with high mean similarity (high $SCI_{\mu}$) but also high variability (high $SCI_{\sigma}$) across space will be down-weighted compared to pairs with similar mean similarity but greater spatial consistency (low $SCI_{\sigma}$). Maximizing $SCI_{prod}$ thus favors grouping channels that are not only spectrally similar on average but also exhibit this similarity consistently across spatial extents, a desirable property for learning features relevant to spatially coherent geospatial phenomena. Channels are then clustered based on these $SCI_{prod}$ scores to form a predefined number of groups, allowing practitioners to balance model capacity and computational budget. Because the spectral structure of the data varies, these groups are typically of unequal size, with each containing a variable number of channels. This grouping allows users to trade off between representational richness and computational efficiency, tailoring the model to the needs of specific applications or datasets.

By construction, the SCI-based grouping strategy reflects spatial autocorrelation: it tends to cluster together bands that show similar spatial behavior across the dataset. This ensures that the resulting groups align well with real-world spatial patterns, allowing TerraMAE to learn representations that are both spectrally and spatially coherent—an essential quality for downstream geospatial tasks.


To assess the effectiveness of our proposed SCI-based adaptive channel grouping, we benchmark its performance against several established alternatives. These comparison methods fall into two broad categories: data-driven clustering techniques and static, domain-informed grouping strategies. Each serves as a point of reference for evaluating the relative merits of SCI in capturing meaningful spectral relationships within hyperspectral data. By contrasting SCI-based grouping with both clustering algorithms that learn from the data and fixed schemes based on wavelength or domain knowledge, we hope to clarify the unique contributions of SCI—especially its ability to leverage spatially consistent spectral similarity for the purpose of more effective representation learning.

\subsubsection{\textbf{Data-Driven Clustering Methods}}
    \begin{itemize}
        \item \textit{k-Means Clustering}: Channels are grouped based on similarity in their statistical properties, computed across the entire training dataset. For each channel, descriptors such as minimum, maximum, mean, standard deviation, dynamic range, coefficient of variation, and self-correlation are extracted. The clustering aims to group spectral bands with similar statistical behavior across the dataset; however, it does not account for spatial context or spatially varying patterns within the imagery.
        
        \item \textit{Hierarchical Agglomerative Clustering (HAC)}: Like k-Means, HAC operates on feature vectors derived from statistical descriptors to explore inherent data structures.
    \end{itemize}
    
    \subsubsection{\textbf{Static Grouping Methods}}
    \begin{itemize}
        \item \textit{VNIR-SWIR (VS) Grouping}: Channels are split into two wavelength-based groups:
        Visible and Near-Infrared (VNIR, approx. 400–1000 nm) and 
        Short-Wave Infrared (SWIR, approx. 1000–2500 nm).
        
        \item \textit{Soil-Reflectance-based (SR) Grouping}: A domain-aware approach dividing channels into five groups based on known soil reflectance patterns, as outlined in the EnMAP science plan~\cite{chabrillat2022enmap}.
    \end{itemize}

The empirical performance of TerraMAE when utilizing SCI-based grouping, compared to these alternative strategies, is detailed in the context of downstream tasks, particularly soil texture prediction (Section \ref{subsub:stp-results}), and in our ablation studies on reconstruction quality (Appendix \ref{appendix:pretrain-exps}). Furthermore, to provide an additional quantitative perspective on the characteristics of the groups formed by these diverse strategies, we analyzed their statistical coherence using silhouette scores (detailed in Appendix \ref{appendix:silhouette}). While general clustering algorithms like k-Means and HAC are well-established \cite{Ran2023, IKOTUN2023178, JAIN2010651}, their specific utility and comparative performance for HSI channel grouping within an MAE framework have not been extensively benchmarked, providing further context for our SCI approach.

\subsection{Enhanced Loss Function - [RQ-1]}
\label{subsec:loss_function}

MAEs have, by convention, relied on pixel-level loss functions—most commonly Mean Squared Error (MSE) or Mean Absolute Error —to evaluate reconstruction quality. These metrics serve well in low-level image reconstruction tasks, especially when the primary goal is to minimize raw intensity differences. However, when applied to spatially organized data such as HSI, these losses often fall short. They are blind to the structural misalignments that may be imperceptible numerically but crucial for preserving spatial coherence \cite{MSE}.

This limitation becomes particularly pronounced in remote sensing applications. HSI data not only captures hundreds of spectrally correlated bands but also reflects complex spatial variations over natural and built environments. When learning representations from such data, relying exclusively on pixel-wise errors may misguide the model—penalizing benign spectral shifts while ignoring spatial boundaries that delineate meaningful land cover units. As a result, critical features such as field edges, crop boundaries, or mineral gradients may be lost in the reconstruction. This undermines the potential of pretraining to support geospatial tasks where preserving both spectral integrity and spatial organization is essential.


To address the limitations of pixel-wise losses in MAEs, we augment the standard Mean Absolute Error loss with two additional components, each designed to preserve a critical dimension of hyperspectral image structure: spatial coherence and spectral fidelity.

To capture spatial coherence, we incorporate the Structural Similarity Index Measure (SSIM), which has been widely used to assess perceptual quality in image processing. Spatial features such as field boundaries, geological formations, and crop textures often manifest as subtle edges, gradients, and textures that are not well-preserved by pixel-level losses alone. SSIM (defined in Equation ~\ref{eq:SSIM}) evaluates local structure through a sliding window that compares luminance, contrast, and structural similarity between corresponding windows in the original and reconstructed spectral bands:

    \begin{equation}
    \operatorname{SSIM}(x,y) = \frac{(2\mu_x\mu_y + c_1)(2\sigma_{xy} + c_2)}{(\mu_x^2 + \mu_y^2 + c_1)(\sigma_x^2 + \sigma_y^2 + c_2)}
    \label{eq:SSIM}
    \end{equation}

    where $\mu_x$, $\mu_y$ are local means, $\sigma_x$, $\sigma_y$ are standard deviations, $\sigma_{xy}$ is covariance, and $c_1$, $c_2$ are stability constants. 
    
    SSIM values range from -1 to 1, with 1 indicating perfect similarity. For its inclusion as a loss component where lower values are better, we define the normalized SSIM loss, $SSIM_N$, as:
    
    \begin{equation}
    \operatorname{SSIM}_N(x,y) = \frac{1 - \operatorname{SSIM}(x,y)}{2}
    \label{eq:SSIM_N}
    \end{equation}
    
    This maps SSIM to a $[0,1]$ range, where 0 signifies perfect reconstruction in terms of structural similarity.

    To address spectral fidelity, we incorporate Spectral Information Divergence (SID), a metric that compares spectral signatures using statistical divergence. This measure is particularly sensitive to differences in both the shape and magnitude of spectral vectors—critical for distinguishing between vegetation types, identifying mineral compositions, and detecting subtle environmental changes. Unlike simpler metrics such as Spectral Angle Mapper \cite{KRUSE1993145}, which focus solely on angular difference, SID captures richer spectral behavior. Though hybrid approaches like SIDSAM \cite{du2004new} have been proposed, we adopt SID due to its balance of discriminative power and implementation efficiency. Given original and reconstructed spectra $x$ and $y$, each with $C$ channels, the SID is defined as:

    \begin{equation}
    \operatorname{SID}(x, y) = \sum_{i=1}^{C} \left( p_i \log\left( \frac{p_i}{q_i} \right) + q_i \log\left( \frac{q_i}{p_i} \right) \right)
    \label{eq:SID}
    \end{equation}
    
    where $p_i$ and $q_i$ are the $i$-th spectral components normalized to sum to one: 
    
    $$
    p_i = \frac{x_i}{\sum_{j=1}^{C} x_j + \epsilon}, \quad q_i = \frac{y_i}{\sum_{j=1}^{C} y_j + \epsilon}
    $$
    
    and $\epsilon$ is a small constant. 
    
    SID values range from 0 (identical spectra) to $\infty$. To scale SID comparably with other loss terms, we normalize it to $[0,1)$ using an exponential transformation:
    \begin{equation}
    \operatorname{SID}_N = 1 - e^{-\alpha \times \operatorname{SID}}
    \label{eq:SID_N}
    \end{equation}
    where $\alpha$ is a positive scaling factor, set to 0.5 in our experiments for balanced mapping and training stability.

Together, these enhancements to the reconstruction loss enable the model to recover hyperspectral images that are not only visually faithful but also structurally and spectrally consistent—characteristics essential for accurate geospatial analysis and interpretation.

Our final loss function is a weighted sum of Mean Absolute Error, normalized SSIM, and normalized SID:

\begin{equation}
  \operatorname{Loss} = \eta \times \operatorname{\text{Mean Absolute Error}} + ~ \lambda \times \operatorname{SSIM}_N + ~ \mu \times \operatorname{SID}_N
  \label{eq:loss_eq}
\end{equation}

While the enhanced loss components increase per-iteration training time by $~2.7\times$ (detailed analysis in Appendix~\ref{sec:loss_computational_analysis}), this overhead is offset by substantially faster convergence and improved downstream task performance (detailed in \ref{subsub:stp-results}).

\subsubsection{\textbf{Loss Weighting Strategy}}
We began by exploring the possibility of learning the loss weights $(\eta, \lambda, \mu)$ dynamically through gradient-based optimization. Although attractive in principle, this approach introduced instability during training. In response, we adopted a linear scheduling strategy that allowed for a more stable and interpretable evolution of loss emphasis over time. Specifically, we initialized training with weights $(1.0, 0.0, 0.0)$—thus focusing solely on pixel-level reconstruction during the early stages—and gradually transitioned toward fixed target weights, chosen through empirical tuning. This trajectory echoes the typical analytical workflow of remote sensing practitioners: broad spatial patterns are first identified, after which attention turns to the finer, more nuanced variations in spectral detail.

To determine an effective final weighting scheme, we conducted a hyperparameter sweep across several representative combinations. Table~\ref{tab:loss_weight_ablation} presents the results of this ablation, reporting reconstruction quality using three standard metrics: Mean Absolute Error, Peak Signal-to-Noise Ratio (PSNR), and Structural Similarity Index (SSIM). Among the configurations tested, the weighting scheme $(\eta = 0.7, \lambda = 0.15, \mu = 0.15)$ emerged as the most effective following 100 epochs of pretraining. It achieved strong and balanced performance across all three metrics.

Although this configuration was selected based on reconstruction accuracy, its benefits extend beyond low-level fidelity. As shown in Table~\ref{tab:grouping_study}, these weights also contribute to improved performance on downstream tasks. This result supports the hypothesis that better spatial and spectral reconstruction yields richer and more transferable representations for geospatial analyses.

\begin{table}[htbp]
\caption{Reconstruction performance with different loss component weights. Higher weights on SSIM and SID improve structural similarity and spectral fidelity at a slight cost to pixel-wise accuracy.}
\label{tab:loss_weight_ablation}
\centering
\begin{small}
\begin{tabular}{c c c | c c c}
\toprule
$\eta$ & $\lambda$ & $\mu$ & MAE $\downarrow$ & PSNR $\uparrow$ & SSIM $\uparrow$ \\
\midrule
1.00 & 0.00 & 0.00 & 0.0418 & 21.08 & 0.3293 \\
0.80 & 0.10 & 0.10 & 0.0139 & 29.08 & 0.6013 \\
\textbf{0.70} & \textbf{0.15} & \textbf{0.15} & \textbf{0.0120} & \textbf{30.18} & \textbf{0.6354} \\
0.40 & 0.30 & 0.30 & 0.0168 & 27.92 & 0.5897 \\
0.30 & 0.50 & 0.20 & 0.0201 & 26.15 & 0.5412 \\
0.30 & 0.20 & 0.50 & 0.0195 & 26.48 & 0.5523 \\
\bottomrule
\end{tabular}
\end{small}
\end{table}

\section{Pretraining Performance and Analysis}

Effective pretraining on hyperspectral data requires careful consideration of how to model its high-dimensional spatial and spectral structure. In this section, we investigate how different architectural strategies, loss functions, and spectral grouping mechanisms contribute to learning geographically meaningful representations from unlabeled HSI data. While masked reconstruction serves as our training objective, our broader goal is to ensure that these learned features are transferable to real-world geospatial prediction tasks. The effectiveness of these design decisions is evaluated through transfer learning experiments in Section~\ref{sec:tle}.

\begin{figure*}[!htbp]
    \centering
    \includegraphics[width=0.9\textwidth, height=7.5cm]{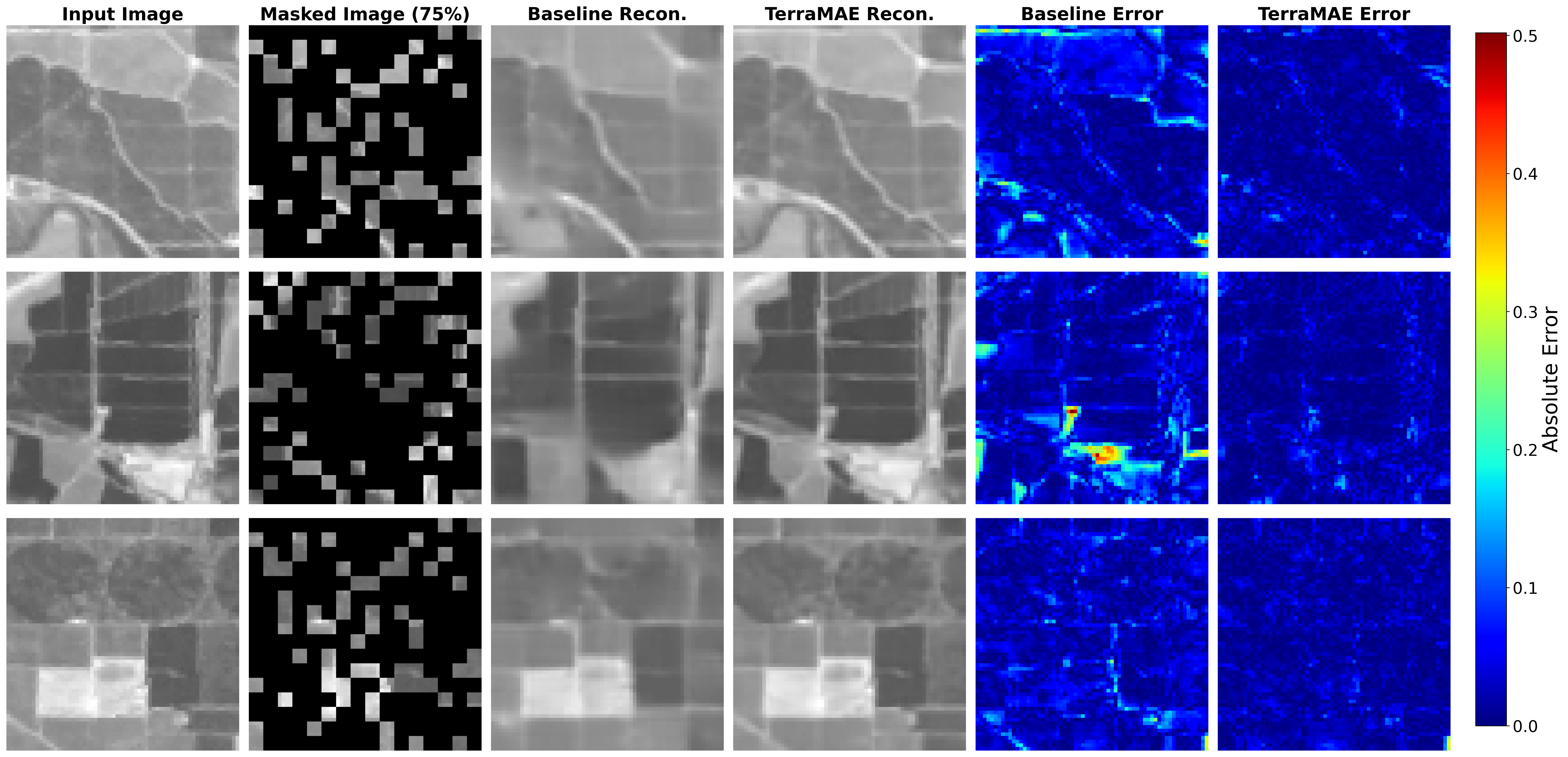}
    \caption{Reconstruction error heatmaps (absolute error) on California test set images, visualized using the spectral channel with the lowest error. TerraMAE demonstrates clearer preservation of spatial structure and lower reconstruction error than the baseline.}
    \Description{Reconstruction Error Heatmap}
    \label{fig:heatmap}
\end{figure*}

\subsection{Experimental Design and Evaluation Framework}

We evaluate TerraMAE's pretraining through multiple lenses, recognizing that reconstruction metrics serve as optimization proxies rather than direct indicators of representation quality. Our experimental framework examines: (1) baseline comparison to validate our design choices, (2) ablation studies on masking ratios and spectral grouping, and (3) geographic generalization through out-of-state reconstruction analysis.

\begin{figure}[!htbp]
    \centering
    \includegraphics[width=\linewidth, height=9cm]{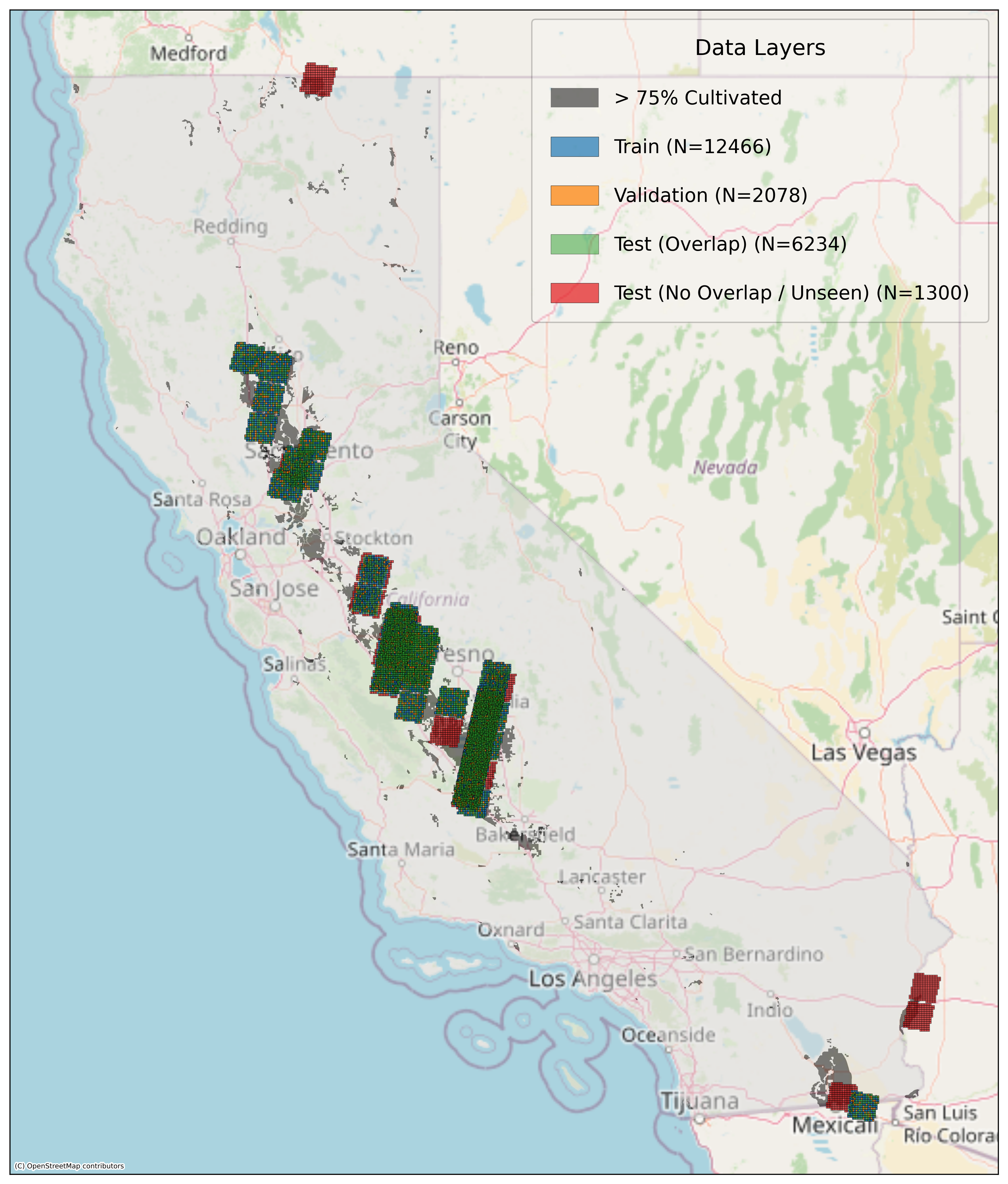}
    \caption{Spatial layout of the EnMAP HSI pretraining dataset in California, demonstrating the strategic focus on agricultural areas based on USDA NASS agricultural strata.}
    \Description{Spatial Layout of Pretraining Dataset covering California.}
    \label{fig:spatial-distribution-CA}
\end{figure}

\subsection{Pretraining Dataset and Spatial Coverage}

Our data collection strategy was guided by the soil texture prediction task (Section~\ref{tle:stp}), which motivated a focus on agricultural regions where high-resolution, spectrally rich hyperspectral data can directly support precision agriculture and soil health monitoring. Consequently, we collected EnMAP (Environmental Mapping and Analysis Program) scenes from California, due to its diverse cropland types concentrated in a geographically compact but heterogeneous region.

To evaluate geographic generalization beyond California, we also collected EnMAP imagery from agricultural regions in Colorado and Kansas. These datasets were excluded from the initial pretraining phase and served two purposes: (1) downstream evaluation in the land cover classification task (Section~\ref{sec:nlcd-exp}), and (2) supplementary analysis of reconstruction performance under both zero-shot and fine-tuned settings (Appendix~\ref{appendix:co-ks-pretrain}). This three-state design tests whether representations learned from California's agricultural diversity can transfer to the high-altitude croplands of Colorado and the Great Plains agriculture of Kansas.

Each GeoTIFF-formatted EnMAP scene collected for pretraining spans approximately 30 km $\times$ 30 km and provides 224 hyperspectral bands covering the 420--2450 nm wavelength range at a 30-meter spatial resolution. The imagery was acquired between April 2022 and August 2024, applying strict quality criteria: only scenes with less than 10\% cloud cover and 0\% snow cover were considered, and mountainous or heavily forested areas were excluded to maintain focus on agricultural landscapes.

To create the final pretraining dataset, we first removed six bands affected by strong water vapor absorption (identified by a uniform value of \texttt{-32768}), resulting in 218 usable spectral channels per tile. We then segmented each large EnMAP scene into non-overlapping 64 $\times$ 64 pixel tiles (2 km $\times$ 2 km ground coverage), yielding a total of 22,078 tiles from California.

We partitioned the California dataset into 12,466 training tiles, 2,078 validation tiles, and 7,534 test tiles. To assess geographic generalization, we split the test set using a spatial overlap criterion. Test Set 2 (TS2) comprises 1,300 tiles from geospatially distinct regions with no spatial overlap with any training or validation tile. The remaining 6,234 test tiles constitute Test Set 1 (TS1). Although TS2 tiles are still within California, they serve to evaluate the model's ability to generalize to unseen regions within the same state. Figure~\ref{fig:spatial-distribution-CA} shows the spatial distribution of all partitions, overlaid with agricultural intensity, illustrating coverage across California's major farming areas.

\begin{table*}[!htbp]
\caption{Image reconstruction results on two test sets. Me. Abs. Err. denotes Mean Absolute Error. TerraMAE employs SCI grouping (5 spectral groups, 75\% masking rate in each group) with the mean absolute error, SSIM, and SID losses, yielding the best overall reconstruction quality. The full table is available in Appendix \ref{appendix:pretrain-exps}.}
\label{table:test-dataset-evalulation-results-short}
\begin{center}
\begin{small}
\begin{tabular}{@{}c c c *{3}{c}|*{3}{c}@{}}
\toprule
\textbf{Setup} & \textbf{Grouping} & \textbf{Loss Function} & \multicolumn{3}{c}{\textbf{Test Set 1}} & \multicolumn{3}{c}{\textbf{Test Set 2 (Unseen)}} \\
\cmidrule(lr){4-6} \cmidrule(lr){7-9}
& & & \textbf{Me. Abs. Err.}~$\downarrow$ & \textbf{PSNR}~$\uparrow$ & \textbf{SSIM}~$\uparrow$ & \textbf{Me. Abs. Err.}~$\downarrow$ & \textbf{PSNR}~$\uparrow$ & \textbf{SSIM}~$\uparrow$ \\
\midrule
Baseline & No Grouping & Me. Abs. Err. 
  & 0.0172 & 27.12 & 0.4227 & 0.0181 & 27.35 & 0.4092 \\
\textbf{TerraMAE} & \textbf{SCI} & \textbf{Me. Abs. Err. + SSIM + SID} 
  & \textbf{0.0047} & \textbf{37.47} & \textbf{0.9112} & \textbf{0.0051} & \textbf{37.55} & \textbf{0.9083} \\
\bottomrule
\end{tabular}
\end{small}
\end{center}
\end{table*}

\subsection{Baseline Model}

To evaluate the effectiveness of our spectral grouping strategy and enhanced loss formulation, we compare TerraMAE against a strong baseline adapted from the original MAE framework. This baseline shares the same ViT-Large backbone—embedding dimension of 1024, 24 transformer blocks, and 16 attention heads—and uses an identical patching structure, but omits spectral grouping and auxiliary loss components. This allows us to isolate the contributions of TerraMAE's architectural and objective function modifications.

\subsection{Implementation Details}
We pretrain TerraMAE for 300 epochs with a 75\% masking ratio using 5 spectral groups determined by our channel grouping algorithms. The model employs standard 2D sin-cos positional embeddings for spatial locations combined with 1D embeddings for channel group indices, enabling both spatial awareness and spectral group distinction. Complete training configurations are provided in Appendix~\ref{appendix:pretrain-config}.

\subsection{Evaluation Metrics}
We evaluate reconstruction performance using Mean Absolute Error, Peak Signal-to-Noise Ratio (PSNR), and Structural Similarity Index (SSIM) \cite{sara2019image, huynh2012accuracy}.

\subsection{Reconstruction Performance and Analysis}
\label{subsec: rpaa}
Table~\ref{table:test-dataset-evalulation-results-short} presents reconstruction performance on the California test sets. TerraMAE, with its SCI grouping and enhanced loss (Mean Absolute Error + SSIM + SID), substantially outperforms the baseline. Notably, performance remains strong on Test Set 2, which contains geographically disjoint tiles not seen during training, indicating that the model learns spatially generalizable features.

Figure~\ref{fig:heatmap} shows TerraMAE's improved reconstruction of spatial detail compared to the baseline—particularly field boundaries—crucial for downstream agricultural applications such as those presented in Section~\ref{sec:tle}.

\section{Transfer Learning Experiments: Evaluating TerraMAE for Geospatial Applications}
\label{sec:tle}

To assess the practical utility of TerraMAE's learned representations for real-world geospatial analysis, we evaluated its performance on three key downstream tasks: \textit{multi-class crop type identification},  \textit{regional land cover classification}, and \textit{soil property prediction}. These downstream tasks were selected because they span a range of spatial scales and geophysical domains—vegetation, land use, and subsurface properties—providing a robust testbed for evaluating the generality and transferability of learned spatial-spectral representations. In each case, TerraMAE's encoder, pretrained in a self-supervised fashion, was frozen, and a lightweight convolutional head was trained to adapt the representations to each task (linear probing). This design enables a direct test of the pretrained encoder's utility, independent of any further tuning. We compared TerraMAE's performance against two baselines: a standard MAE (without the architectural and algorithmic enhancements in TerraMAE) and a ResNet-50 model trained from scratch for each task. The inclusion of ResNet-50, a widely used, deeper architecture, provides a meaningful point of comparison for assessing representational strength versus TerraMAE.


\subsubsection{\textbf{Downstream Model Architecture and Training Setup}} For all downstream tasks, we employed a lightweight convolutional decoder to map frozen encoder features to task-specific outputs. The architecture consists of alternating convolutional and transposed convolutional layers with batch normalization and Rectified Linear Unit (ReLU) activation functions, progressively upscaling spatial features to $64 \times 64$ resolution. We applied task-specific loss functions: cross-entropy loss for classification tasks (crop type and land cover mapping), and mean absolute error for soil texture regression. The decoder networks were optimized using stochastic gradient descent with a batch size of 32 and learning rates around $10^{-4}$ for up to 150 epochs, with hyperparameters tuned for each specific task.


\subsection{Soil Texture Prediction}
\label{tle:stp}

Soil texture, which is determined by the relative proportions of sand, silt, and clay, plays a foundational role in governing hydrological and nutrient-related soil behaviors, including water retention, infiltration rates, and nutrient availability. Generating accurate, spatially-resolved maps of soil texture is therefore vital for a range of land management practices, particularly those underpinning precision agriculture, such as targeted irrigation and fertilization strategies. HSI, with its capacity to capture subtle spectral variations tied to surface mineralogy and moisture, offers a compelling modality for this task. Because the prediction of soil texture served as a central motivation in the design of our EnMAP dataset, it provides a natural and rigorous benchmark for assessing the effectiveness of TerraMAE's pretraining in extracting transferable spatial-spectral representations.

\begin{figure}[!htbp]
\centering
\includegraphics[width=\linewidth]{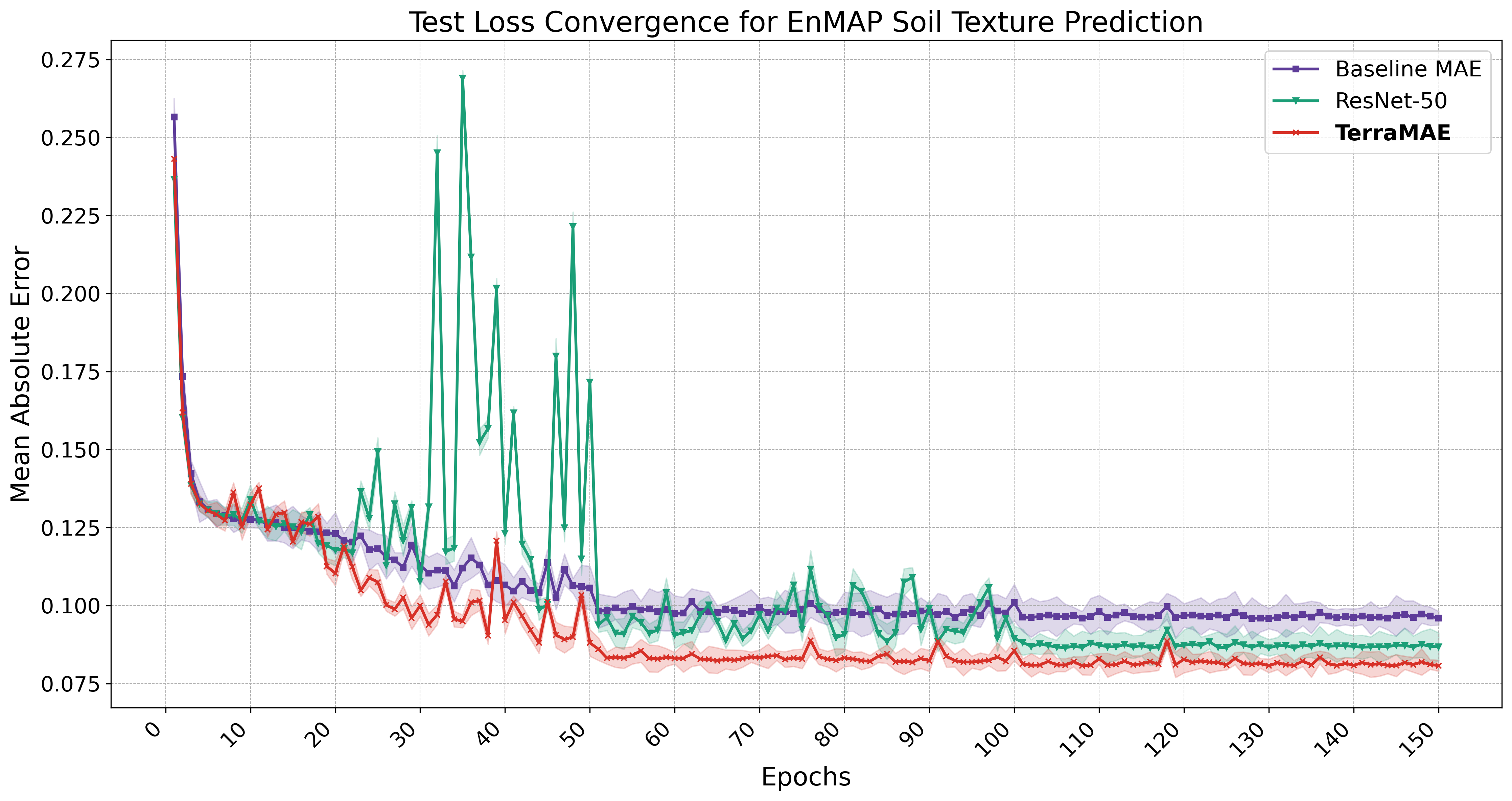}
\caption{Test loss convergence over 150 epochs for soil texture prediction using different encoder initializations. TerraMAE achieves lower error and smoother convergence.}
\Description{Soil Texture Prediction Test Loss Convergence}
\label{fig:convergence_soil}
\end{figure}

\subsubsection{\textbf{Dataset and Methodology}} We used the POLARIS (Probabilistic Reanalysis of Soil Survey Information for the U.S.) dataset \cite{CHANEY201654}, which provides 30m resolution estimates of sand, silt, and clay across the U.S. For California, we extracted per-pixel probability estimates of sand and silt as supervision targets. Using features from our pretrained EnMAP encoders, downstream models were trained to predict $64 \times 64 \times 2$ maps for sand and silt; clay was inferred to ensure the three fractions sum to 100\%. Performance was measured using Mean Absolute Error against POLARIS references.

\begin{table}[!htbp]
\caption{EnMAP-POLARIS Soil Texture Prediction Results (Mean Absolute Error $\pm $ Standard Deviation over 5 runs).}
\label{tab:stp_performance_updated}
\begin{center}
\begin{small}
\begin{tabular}{l c}
\toprule
\textbf{Method} & \textbf{Mean Absolute Error $\pm$ Std. Dev. $\downarrow$} \\
\midrule
ResNet-50 & 0.0863 $\pm$ 0.0032 \\
Baseline MAE + CNN & 0.0956 $\pm$ 0.0041 \\
\textbf{TerraMAE + CNN} & \textbf{0.0806 $\pm$ 0.0023} \\
\bottomrule
\end{tabular}
\end{small}
\end{center}
\end{table}


\subsubsection{\textbf{Results and Discussion}}
\label{subsub:stp-results}
Summarized in Table~\ref{tab:stp_performance_updated}, TerraMAE + CNN achieved the lowest Mean Absolute Error, indicating the highest prediction accuracy. Specifically, the CNN (with 220k parameters) with TerraMAE yields a 6.60\% relative improvement over the supervised ResNet-50 baseline (with around 25M parameters) and a 15.69\% improvement over the Baseline MAE + CNN model. This demonstrates that effective pretraining on HSI data can compensate for architectural simplicity—our pretrained representations enable a compact decoder to outperform a deeper supervised network trained from scratch. These results are notable, particularly given the difficulty of regressing soil texture components from remote sensing data. TerraMAE's superior performance suggests its HSI-specific pretraining enables more discriminative spectral-spatial feature extraction relevant to soil properties. 
Figure~\ref{fig:convergence_soil} shows that TerraMAE not only reaches the lowest Mean Absolute Error but also converges more smoothly—important for geospatial regression tasks where training instability can impact map quality.

\begin{table}[!htbp]
\caption{
Effect of Channel Grouping Strategies in Pretraining on Soil Texture Prediction. 
\textbf{Me. Abs. Err.} denotes Mean Absolute Error. 
\textbf{SR} refers to the Spectral Reflectance strategy, 
\textbf{HAC} to Hierarchical Agglomerative Clustering, and 
\textbf{SCI} to Spectral Comparison Index. 
\textbf{SCI achieves the best performance among all strategies using both the default and enhanced loss functions.}
}
\label{tab:grouping_study}
\begin{center}
\begin{small}
\begin{tabular}{l l c}
\toprule
\textbf{Grouping Strategy} & \textbf{Loss Function} & \textbf{Me. Abs. Err. $\downarrow$} \\
\midrule
None & Me. Abs. Err. & 0.1027 \\
None & Me. Abs. Err. + SSIM + SID & 0.0956 \\
SR          & Me. Abs. Err. & 0.0938 \\
SR          & Me. Abs. Err. + SSIM + SID & 0.0907 \\
k-Means     & Me. Abs. Err. & 0.0879 \\
k-Means     & Me. Abs. Err. + SSIM + SID & 0.0843 \\
HAC         & Me. Abs. Err. & 0.0855 \\
HAC         & Me. Abs. Err. + SSIM + SID & 0.0822 \\
\textbf{SCI} & \textbf{Me. Abs. Err.} & \textbf{0.0828} \\
\textbf{SCI} & \textbf{Me. Abs. Err. + SSIM + SID} & \textbf{0.0806} \\
\bottomrule
\end{tabular}
\end{small}
\end{center}
\end{table}

To isolate the impact of grouping strategies in pretraining, we evaluated downstream soil texture prediction using a fixed CNN head across all models. As shown in Table~\ref{tab:grouping_study}, SCI grouping yielded the best performance, confirming that spectral similarity-based grouping enhances transferability. Improvements across all alternatives suggest that grouping-aware pretraining benefits geospatial prediction beyond reconstruction.

\subsection{Multi-Class Crop Type Identification}

Prompt identification of crop types is crucial for applications in precision agriculture, yield forecasting, land-use planning, and environmental monitoring. HSI, with its detailed spectral signatures, offers a distinct advantage over traditional multispectral data for differentiating crop types that may exhibit subtle spectral variations, thereby enabling more nuanced agricultural analysis.

\subsubsection{\textbf{Dataset and Methodology}} For this study, we utilized the Cropland Data Layer (CDL) for California (2022–2023), which provides annual land cover classifications at a 30-meter spatial resolution. CDL is a widely validated benchmark for agricultural land cover analysis and model evaluation in the U.S.

Our EnMAP HSI collection was strategically focused on agricultural regions within California (CA), driven by the data requirements of our primary downstream task—soil texture prediction (Section~\ref{tle:stp}). This geographic constraint inherently influenced the distribution and variety of CDL crop classes represented in our dataset, resulting in a focused yet representative sample of crop types relevant to CA's agricultural landscape.

\begin{table}[!htbp]
\caption{EnMAP-CDL Dataset Class Distribution (CA)}
\label{tab:class_dist}
\begin{center}
\small
\begin{tabular}{l c}
\toprule
\textbf{Class Name} & \textbf{Pixel Share (\%)} \\
\midrule
Almonds & 18.3 \\
Grass/Pasture & 6.6 \\
Pistachios & 10.5 \\
Fallow/Idle Cropland & 11.3 \\
Alfalfa & 4.9 \\
Minor Crop Types (aggregated) & 48.3 \\
\bottomrule
\end{tabular}
\end{center}
\end{table}

To support crop classification experiments, we selected CDL classes that had a significant presence within the subset of EnMAP image tiles overlapping with CDL annotations. Specifically, we curated six crop categories: \textit{Almonds}, \textit{Grass/Pasture}, \textit{Pistachios}, \textit{Fallow/Idle Cropland}, \textit{Alfalfa}, and a consolidated \textit{Minor Crop Types} class. The \textit{Minor Crop Types} category aggregates CDL crop types that individually accounted for less than 5\% of the total pixel count in the aligned EnMAP-CDL dataset—a practical strategy often employed in remote sensing classification tasks to address class imbalance and ensure robust model training across all categories.

The class distribution within the aligned EnMAP-CDL dataset, based on pixel frequency, is summarized in Table~\ref{tab:class_dist}. \textit{Minor Crop Types} account for the largest share of labeled pixels, while \textit{Alfalfa} and \textit{Grass/Pasture} are underrepresented. This class imbalance has implications for model performance, particularly in per-class evaluation metrics. To mitigate the effects of this skewed distribution, we applied a class-weighted loss during CNN training to emphasize minority categories. For both training and evaluation, $64 \times 64$ pixel tiles aligned with EnMAP HSI coverage were used, with an 80/20 train–test split.

\subsubsection{\textbf{Results and Discussion}}
\label{subsubsec:cdl-results}
Table \ref{tab:cdl_performance_terraMAE} details the per-class and mean performance for the multi-class crop type identification task. TerraMAE + CNN consistently achieved the highest mean Top-1 Accuracy and mean Intersection over Union (IoU). This underscores the benefit of TerraMAE's HSI-specific pretraining for this challenging fine-grained classification task.

While TerraMAE generally outperforms baselines for individual crop types, and the aggregated \textit{Minor Crop Types} class performs well, performance for certain specific classes like \textit{Pistachios} and the minority classes \textit{Alfalfa} and \textit{Grass/Pasture} remains comparatively limited. These class-dependent variations are well-documented in remote sensing-based agricultural mapping. For instance, the Prithvi-100M foundation model reported similarly uneven per-class accuracies when fine-tuned on CDL labels (\textit{e.g.}, Alfalfa IoU of 0.3084) \cite{jakubik2023foundationmodelsgeneralistgeospatial}. Such findings underscore the inherent difficulty of fine-grained crop classification.


\begin{table}[htbp]
\caption{Crop Type Identification on CDL data (Top 6 classes in California). Best results for each class are in bold.}
\label{tab:cdl_performance_terraMAE}
\begin{center}
\resizebox{\columnwidth}{!}{ 
\begin{tabular}{l c c c c c c}
\toprule
\textbf{Class Name} & \multicolumn{2}{c}{\textbf{ResNet-50}} & \multicolumn{2}{c}{\textbf{Baseline MAE + CNN}} & \multicolumn{2}{c}{\textbf{TerraMAE + CNN}} \\
\cmidrule(lr){2-3} \cmidrule(lr){4-5} \cmidrule(lr){6-7}
 & \textbf{Acc (\%) $\uparrow$} & \textbf{IoU (\%) $\uparrow$} & \textbf{Acc (\%) $\uparrow$} & \textbf{IoU (\%) $\uparrow$} & \textbf{Acc (\%) $\uparrow$} & \textbf{IoU (\%) $\uparrow$} \\
\midrule
Almonds                & 54.48 & 27.32 & 51.48 & 25.14 & \textbf{56.41} & \textbf{30.59} \\
Grass/Pasture          & 60.46 & 35.68 & 57.53 & 33.77 & \textbf{62.43} & \textbf{37.22} \\
Pistachios             & 31.44 & 16.50 & 34.89 & 18.59 & \textbf{37.26} & \textbf{20.80} \\
Fallow/Idle Cropland   & 35.38 & 18.70 & 34.92 & 17.84 & \textbf{42.33} & \textbf{23.50} \\
Alfalfa                & 34.21 & 17.90 & 39.73 & 21.96 & \textbf{42.12} & \textbf{23.10} \\
Minor Crop Types                  & 83.49 & 50.07 & 82.68 & 49.38 & \textbf{86.21} & \textbf{53.27} \\
\midrule
\textbf{Mean}          & 49.91 & 27.03 & 50.87 & 27.78 & \textbf{54.79} & \textbf{31.41} \\
\bottomrule
\end{tabular}
}
\end{center}
\end{table}
\subsection{Land Cover Classification}
\label{sec:nlcd-exp}
Accurate land cover mapping is a cornerstone of geospatial analysis, providing essential baseline information for urban and regional planning, environmental monitoring, natural resource management, and climate change impact assessment. Improving the accuracy and detail of such maps using advanced remote sensing techniques like HSI remains a critical research area.

\begin{table}[htbp]
\caption{Class Distribution in the NLCD Dataset (Study Region: California, Colorado, and Kansas)}
\label{tab:class_dist_nlcd}
\begin{center}
\small
\begin{tabular}{l c}
\toprule
\textbf{Land Cover Name} & \textbf{Pixel Share (\%)} \\
\midrule
Shrub/Scrub & 25.44 \\
Minor Land Cover Types & 20.03 \\
Herbaceous & 19.14 \\
Evergreen Forest/Mixed Forest & 18.20 \\
Cultivated Crops & 17.19 \\
\bottomrule
\end{tabular}
\end{center}
\end{table}

\subsubsection{\textbf{Dataset and Methodology}}
For this task, we utilized the 2021 National Land Cover Database (NLCD) for California, Colorado, and Kansas, focusing on extracting land cover classes that spatially correspond to our EnMAP HSI study areas. The NLCD provides consistent, high-resolution (30-meter) land cover information across the United States and is widely used in geospatial analysis as a reliable reference for land cover classification tasks. In our work, it serves as a standardized benchmark for evaluating model performance across diverse land cover types.

Based on the land cover types present within our EnMAP HSI collection and their spatial distribution, we defined five primary NLCD-based classes for evaluation: \textit{Shrub/Scrub}, \textit{Herbaceous}, \textit{Evergreen Forest/Mixed Forest}, \textit{Cultivated Crops}, and a consolidated \textit{Minor Land Cover Types} category. The \textit{Minor Land Cover Types} class aggregates NLCD types that individually constituted less than 10\% of the total pixel count and were not central to the focus of this analysis. This aggregation strategy ensures that all explicitly modeled classes have sufficient representation for training and evaluation.

The pixel distribution across these classes in our test set is summarized in Table~\ref{tab:class_dist_nlcd}. While the distribution reflects a reasonably diverse set of land cover types, it remains moderately imbalanced, with \textit{Shrub/Scrub} being the most prevalent and \textit{Cultivated Crops} among the least frequent specific categories. As with the CDL task, we employed a class-weighted loss function  within the downstream CNN classification heads, allowing underrepresented classes to contribute proportionally to the learning objective. For all experiments, we used an $80/20$ train-test split of $64 \times 64$ pixel tiles that were spatially aligned with the EnMAP HSI coverage.

\begin{table}[htbp]
\caption{Land Cover Classification on NLCD (Top 5 classes in California, Colorado, \& Kansas). Best results in bold.}
\label{tab:nlcd_performance_terraMAE}
\begin{center}
\resizebox{\columnwidth}{!}{%
\begin{tabular}{l c c c c c c}
\toprule
\textbf{Class Name} & \multicolumn{2}{c}{\textbf{ResNet-50}} & \multicolumn{2}{c}{\textbf{Baseline MAE + CNN}} & \multicolumn{2}{c}{\textbf{TerraMAE + CNN}} \\
\cmidrule(lr){2-3} \cmidrule(lr){4-5} \cmidrule(lr){6-7}
& \textbf{Acc (\%) $\uparrow$} & \textbf{IoU (\%) $\uparrow$} & \textbf{Acc (\%) $\uparrow$} & \textbf{IoU (\%) $\uparrow$} & \textbf{Acc (\%) $\uparrow$} & \textbf{IoU (\%) $\uparrow$} \\
\midrule
Shrub/Scrub & 64.15 & 41.33 & 65.07 & 42.14 & \textbf{74.33} & \textbf{46.19} \\
Evergreen Forest/Mixed Forest & 79.41 & 58.30 & 84.66 & 63.20 & \textbf{90.05} & \textbf{66.72} \\
Herbaceous & 60.47 & 39.50 & 59.49 & 38.00 & \textbf{68.74} & \textbf{43.55} \\
Minor Land Cover Types & 88.74 & 68.10 & 87.23 & 67.05 & \textbf{93.37} & \textbf{71.55} \\
Cultivated Crops & 87.68 & 65.26 & 85.53 & 63.91 & \textbf{92.61} & \textbf{69.04} \\
\midrule
\textbf{Mean} & 76.09 & 54.50 & 76.40 & 54.86 & \textbf{83.82} & \textbf{59.41} \\
\bottomrule
\end{tabular}%
}
\end{center}
\end{table}

\subsubsection{\textbf{Results and Discussion}}
The performance of TerraMAE and baseline models on the NLCD land cover classification task is presented in Table \ref{tab:nlcd_performance_terraMAE}. TerraMAE + CNN achieved the highest mean Top-1 Accuracy and mean IoU, improving significantly over both the ResNet-50 baseline and the Baseline MAE + CNN and demonstrating the substantial benefit of its HSI-specific pretraining for this land cover mapping task.

Across all three downstream tasks, TerraMAE demonstrates notable gains in classification accuracy and regression precision over both baseline MAE and ResNet-50, highlighting the effectiveness of our pretraining strategy for high-dimensional hyperspectral data. Notably, unlike ResNet-50, TerraMAE's encoder requires no fine-tuning; only a lightweight classifier is trained, underscoring the value of TerraMAE as a foundational model for HSI.




\section{Conclusion}
In this work, we developed TerraMAE, a self-supervised model designed to learn generalizable spatial-spectral representations for downstream geospatial tasks.
\textbf{[RQ-1]} We targeted a reconstruction objective that jointly preserves spatial structure and spectral fidelity in HSI. Our composite loss function integrates pixel-level error with measures of structural similarity and spectral divergence. This allowed the pretraining objective to account for both the local spatial patterns and the inter-band spectral relationships that are essential to geoscientific interpretation.
\textbf{[RQ-2]} We proposed an adaptive channel grouping method driven by spatial-spectral coherence, rather than fixed wavelength partitions. This enabled TerraMAE to learn from the full range of spectral variability while respecting the statistical dependencies that arise in natural imagery.
Across a suite of evaluation benchmarks—including reconstruction fidelity and performance on real-world tasks such as crop classification, land cover mapping, and soil texture prediction—TerraMAE consistently outperformed both a standard MAE baseline and a supervised ResNet-50. These results suggest that TerraMAE can provide powerful representations that transfer effectively across geospatial domains.
Future work will examine how temporal dynamics can be incorporated into the TerraMAE framework.


\begin{acks}
This research was supported by the National Science Foundation (1931363, 2312319), the National Institute of Food Agriculture \\
(COL014021223), an NSF/NIFA Artificial Intelligence Institutes AI-LEAF [2023-03616] and the Clare Booth Luce Professorship.
\end{acks}





\bibliographystyle{ACM-Reference-Format}
\bibliography{sample-base}


\clearpage

\appendix

\section{Appendix}

\subsection{Runtime and Convergence Impact of Spatial-Spectral Loss Components}
\label{sec:loss_computational_analysis}

\begin{figure}[!htbp]
    \centering
    \includegraphics[width=\linewidth]{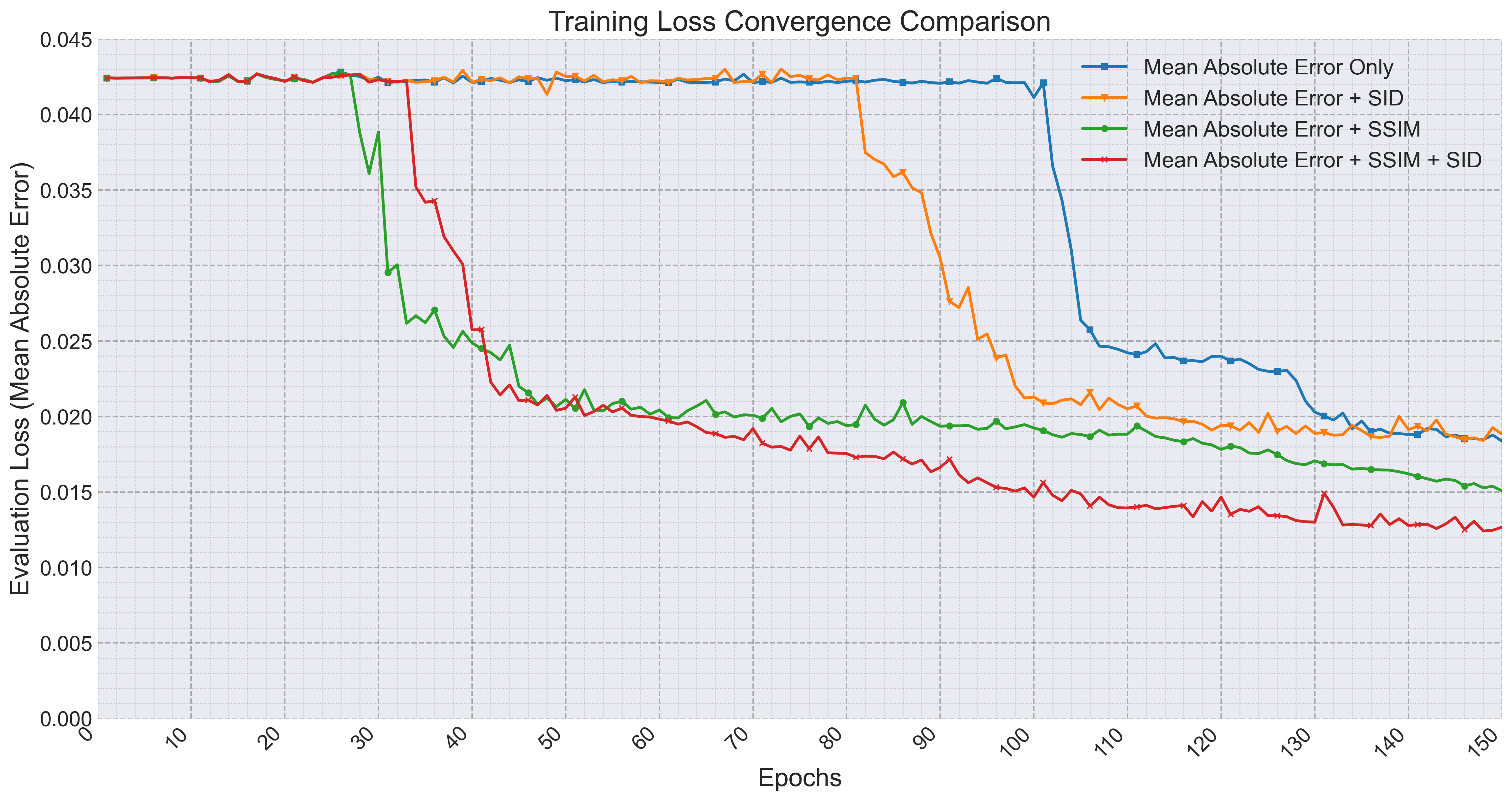}
    \caption{Convergence of the evaluation error on the pretraining validation set over 150 epochs for different loss function configurations: The TerraMAE configuration (Mean Absolute Error + SSIM + SID) demonstrates the most effective convergence to the lowest error.}
    \Description{TerraMAE pretraining convergence of error loss configurations}
    \label{fig:loss_comparison}
\end{figure}

We assess the computational impact of integrating SSIM and SID into TerraMAE's loss function and their effect on convergence during pretraining. 


All experiments were conducted on a single NVIDIA A100 GPU (batch size: 32), with timings averaged over 150 iterations. Table~\ref{tab:loss_computation_time_revised} summarizes per-batch runtime across loss configurations. Incorporating SID introduces only modest overhead ($\sim$14\%), due to its lightweight spectral operations. SSIM, which uses sliding-window convolutions, increases batch time more substantially ($\sim\!2.6\times$ vs. Mean Absolute Error-only). The full TerraMAE loss (Mean Absolute Error + SSIM + SID) results in a total overhead of $2.7\times$.

\begin{table}[H]
\caption{Per-batch runtime for different loss configurations.}
\label{tab:loss_computation_time_revised}
\centering
\resizebox{\columnwidth}{!}{%
\begin{tabular}{l c c c}
\toprule
\textbf{Loss Configuration} & \textbf{Forward (s)} & \textbf{Backward (s)} & \textbf{Total (s) per Batch} \\
\midrule
MAE Only & 1.26 & 1.76 & 3.02 \\
MAE + SID & 1.41 & 2.03 & 3.44 \\
MAE + SSIM & 2.84 & 4.92 & 7.76 \\
\textbf{MAE + SSIM + SID (TerraMAE Loss)} & \textbf{2.99} & \textbf{5.19} & \textbf{8.18} \\
\bottomrule
\end{tabular}%
}
\end{table}

Despite the additional cost, Figure~\ref{fig:loss_comparison} shows that the combined loss significantly improves convergence behavior, reaching the lowest final error and exhibiting stable training dynamics. These results support the inclusion of SSIM and SID in TerraMAE's objective: the added overhead is justified by improved feature learning and reconstruction fidelity, critical for downstream geospatial tasks.

\subsection{Silhouette Scores of Channel Grouping Strategies}
\label{appendix:silhouette}

We evaluated the statistical coherence of our channel grouping strategies using silhouette scores computed on normalized spectral features (mean reflectance, standard deviation, dynamic range, coefficient of variation, and self-correlation). The silhouette score $s = \frac{b-a}{\max(a,b)}$ measures intra-cluster cohesion ($a$) versus inter-cluster separation ($b$), ranging from -1 to 1, with higher values indicating more compact and well-separated groups.

\begin{table}[H]
\centering
\caption{Silhouette scores for channel grouping strategies}
\label{tab:silhouette-scores}
\small
\begin{tabular}{lcc}
\toprule
\textbf{Strategy} & \textbf{Score} & \textbf{Optimization Basis} \\
\midrule
KMeans (5 groups) & 0.557 & Statistical clustering \\
HAC (5 groups) & 0.526 & Statistical clustering \\
SCI (5 groups) & 0.412 & Spatial-spectral similarity \\
VNIR-SWIR (2 groups) & 0.297 & Physical sensor ranges \\
Soil-Reflectance (5 groups) & 0.143 & Domain knowledge \\
\bottomrule
\end{tabular}
\end{table}

Statistical clustering methods (KMeans, HAC) achieved the highest scores as expected, since they directly optimize the feature space used for evaluation. Notably, our SCI method achieved a competitive score (0.412, Table ~\ref{tab:silhouette-scores}) despite optimizing for spatial-spectral patterns rather than statistical features, suggesting it captures both spatial coherence and statistical structure.
Domain-based strategies (VNIR-SWIR, Soil-Reflectance) showed lower scores, reflecting their focus on physical properties rather than statistical compactness. However, silhouette scores alone do not determine grouping efficacy for downstream tasks—the true utility emerges from learned representation (Section~\ref{sec:tle}) and reconstruction quality (Section ~\ref{subsec: rpaa}).

\begin{table*}[!htbp]
\caption{Image reconstruction results on two test sets. TerraMAE (row 24) yields the best overall reconstruction quality.}
\label{table:test-dataset-evalulation-results}
\begin{center}
\begin{scriptsize}
\begin{tabular}{@{}c c c *{3}{c}|*{3}{c}@{}}
\toprule
\textbf{Identifier} & \textbf{Grouping} & \textbf{Loss Function} & \multicolumn{3}{c}{\textbf{Test Set 1}} & \multicolumn{3}{c}{\textbf{Test Set 2 (Unseen)}} \\
\cmidrule(lr){4-6} \cmidrule(lr){7-9}
& & & \textbf{Me. Abs. Err.}~$\downarrow$ & \textbf{PSNR}~$\uparrow$ & \textbf{SSIM}~$\uparrow$ & \textbf{Me. Abs. Err.}~$\downarrow$ & \textbf{PSNR}~$\uparrow$ & \textbf{SSIM}~$\uparrow$ \\
\midrule
1 (Baseline) & No Grouping & Me. Abs. Err. 
  & 0.0172 & 27.12 & 0.4227 & 0.0181 & 27.35 & 0.4092 \\
2 & No Grouping & Me. Abs. Err. + SSIM 
  & 0.0097 & 31.39 & 0.7395 & 0.0108 & 31.42 & 0.7169 \\
3 & No Grouping & Me. Abs. Err. + SID 
  & 0.0193 & 26.44 & 0.3556 & 0.0204 & 25.82 & 0.3408 \\
4 & No Grouping & Me. Abs. Err. + SSIM + SID 
  & 0.0118 & 30.44 & 0.7491 & 0.0134 & 30.12 & 0.7246 \\
\midrule
5 & VNIR-SWIR (VS) & Me. Abs. Err. 
  & 0.0103 & 31.37 & 0.7006 & 0.0110 & 31.53 & 0.6859 \\
6 & Soil-Reflectance (SR) & Me. Abs. Err. 
  & 0.0101 & 31.74 & 0.7072 & 0.0106 & 31.59 & 0.6966 \\
7 & kMeans & Me. Abs. Err. 
  & 0.0097 & 31.99 & 0.7287 & 0.0103 & 32.19 & 0.7199 \\
8 & HAC & Me. Abs. Err. 
  & 0.0082 & 33.16 & 0.7939 & 0.0086 & 32.33 & 0.7914 \\
9 & SCI & Me. Abs. Err. 
  & 0.0078 & 33.66 & 0.7903 & 0.0084 & 33.59 & 0.7876 \\
\midrule
10 & VS & Me. Abs. Err. + SSIM 
  & 0.0098 & 31.76 & 0.7316 & 0.0105 & 31.98 & 0.7241 \\
11 & VS & Me. Abs. Err. + SID 
  & 0.0137 & 30.11 & 0.6952 & 0.0138 & 25.90 & 0.6806 \\
12 & VS & Me. Abs. Err. + SSIM + SID 
  & 0.0087 & 32.77 & 0.8406 & 0.0095 & 33.04 & 0.8288 \\
\midrule
13 & SR & Me. Abs. Err. + SSIM 
  & 0.0094 & 32.34 & 0.7813 & 0.0105 & 32.09 & 0.7762 \\
14 & SR & Me. Abs. Err. + SID 
  & 0.0115 & 31.52 & 0.7593 & 0.0126 & 31.36 & 0.7520 \\
15 & SR & Me. Abs. Err. + SSIM + SID 
  & 0.0086 & 33.44 & 0.8429 & 0.0095 & 33.21 & 0.8368 \\
\midrule
16 & kMeans & Me. Abs. Err. + SSIM 
  & 0.0099 & 31.79 & 0.7208 & 0.0105 & 32.09 & 0.7189 \\
17 & kMeans & Me. Abs. Err. + SID 
  & 0.0087 & 32.81 & 0.7946 & 0.0093 & 33.04 & 0.7944 \\
18 & kMeans & Me. Abs. Err. + SSIM + SID 
  & 0.0085 & 33.52 & 0.8548 & 0.0092 & 33.42 & 0.8520 \\
\midrule
19 & HAC & Me. Abs. Err. + SSIM 
  & 0.0057 & 36.07 & 0.8846 & 0.0063 & 36.09 & 0.8820 \\
20 & HAC & Me. Abs. Err. + SID 
  & 0.0084 & 33.03 & 0.8067 & 0.0089 & 33.11 & 0.8073 \\
21 & HAC & Me. Abs. Err. + SSIM + SID 
  & 0.0084 & 33.62 & 0.8563 & 0.0091 & 33.55 & 0.8558 \\
\midrule
22 & SCI & {Me. Abs. Err. + SSIM} 
  & 0.0083 & 33.78 & 0.8576 & 0.0088 & 33.66 & 0.8559 \\
23 & SCI & Me. Abs. Err. + SID 
  & 0.0098 & 32.06 & 0.7638 & 0.0108 & 31.91 & 0.7578 \\
\textbf{24 (TerraMAE)} & \textbf{SCI} & \textbf{Me. Abs. Err. + SSIM + SID} 
  & \textbf{0.0047} & \textbf{37.47} & \textbf{0.9112} & \textbf{0.0051} & \textbf{37.55} & \textbf{0.9083} \\
\bottomrule
\end{tabular}
\end{scriptsize}
\end{center}
\end{table*}



\subsection{Pretraining Configurations}
\label{appendix:pretrain-config}
\begin{itemize}
    \item \textbf{Mask ratio}: 75\%
    \item \textbf{Patch size}: 4 $\times$ 4
    \item \textbf{Input size}: 64 $\times$ 64 $\times$ 218
    \item \textbf{\# of Spectral groups}: 5
    \item \textbf{Training epochs}: 300
    \item \textbf{Batch size}: 32
    \item \textbf{Optimizer}: AdamW ($\beta_1$=0.9, $\beta_2$=0.95)
    \item \textbf{Learning rate}: $10^{-3}$ with cosine annealing scheme
    \item \textbf{Weight decay}: 0.05
    \item \textbf{Hardware}: NVIDIA A100 GPU
    \item \textbf{Training time}: 12--15 hours
\end{itemize}

\subsection{Complete Pretraining Reconstruction Results on California Test Sets}
\label{appendix:pretrain-exps}

The comprehensive results in Table ~\ref{table:test-dataset-evalulation-results} reveal three critical insights:
\begin{itemize}
    \item \textit{Channel grouping consistently improves reconstruction}: All grouping strategies (rows 5-24) outperform the baseline no-grouping approach (rows 1-4), with improvements ranging from 20\% to 73\% in mean absolute error.
    \item \textit{Multi-objective loss functions enhance quality}: Combining Mean Absolute Error with structural (SSIM) and spectral (SID) losses improves both spatial coherence and spectral fidelity. The triple combination consistently yields the best results within each grouping strategy.
    \item \textit{SCI grouping achieves superior performance}: Our SCI method shows strong performance even with Mean Absolute Error alone (row 9), and when combined with SSIM and SID losses (row 24), achieves the best overall results.
\end{itemize}


\subsection{Geographic Generalization to CO and KS}
\label{appendix:co-ks-pretrain}


\begin{table}[H]
\caption{Reconstruction on CO and KS datasets. TS1 and TS2 are intra-state test sets, both unseen during ZS.}
\label{tab:appendix_co_ks_reconstruction}
\centering
\scriptsize 
\begin{tabular}{@{}llcclll@{}}
\toprule
\textbf{Model} & \textbf{Setting} & \textbf{State} & \textbf{Test Set} & \textbf{Me. Abs. Err. $\downarrow$} & \textbf{PSNR $\uparrow$} & \textbf{SSIM $\uparrow$} \\
\midrule
TerraMAE & ZS & CO & TS1 & 0.0144 & 26.25 & 0.8630 \\
         &    &    & TS2 & 0.0130 & 26.56 & 0.8670 \\
\cmidrule(lr){2-7}
         & FT & CO & TS1 & 0.0062 & 33.66 & 0.8987 \\
         &         &    & TS2 & 0.0053 & 34.88 & 0.8967 \\
\cmidrule(lr){2-7}
         & ZS & KS & TS1 & 0.0144 & 26.25 & 0.8630 \\ 
         &    &    & TS2 & 0.0130 & 26.56 & 0.8670 \\ 
\midrule
Baseline MAE & ZS & CO & TS1 & 0.0559 & 19.10 & 0.3388 \\
            &    &    & TS2 & 0.0564 & 19.31 & 0.3511 \\
\cmidrule(lr){2-7}
            & FT & CO & TS1 & 0.0218 & 23.85 & 0.3635 \\
            &         &    & TS2 & 0.0182 & 25.08 & 0.3669 \\
\cmidrule(lr){2-7}
            & ZS & KS & TS1 & 0.0424 & 21.58 & 0.3449 \\
            &    &    & TS2 & 0.0410 & 21.97 & 0.3432 \\
\bottomrule
\end{tabular}
\end{table}

To supplement downstream classification, we assessed TerraMAE's reconstruction generalization to EnMAP data from Colorado (CO) and Kansas (KS), both excluded from California (CA) pretraining, under two settings:

\begin{itemize}
    \item \textbf{Zero-Shot (ZS):} Direct application of CA-pretrained TerraMAE and baseline MAE on CO and KS.
    \item \textbf{Fine-tuning (FT, CO only):} 100-epoch adaptation on 30\% of CO training tiles.
    \item \textbf{Dataset Sizes:} CO: 814 train / 271 val / 2,741 test tiles; KS: 2,377 test tiles.
\end{itemize}

From Table ~\ref{tab:appendix_co_ks_reconstruction}, TerraMAE significantly outperforms baseline MAE across all CO/KS subsets in ZS settings. Fine-tuning on 30\% of CO data yields a >50\% Mean Absolute Error reduction and +8 dB PSNR gain, further validating TerraMAE's robustness for cross-regional generalization.

\subsection{Ablation Studies}


\begin{table}[H]
\caption{Varying mask ratios on California Test Set 2.}
\label{tab:masking-ratio-comparison}
\begin{center}
\scriptsize
\begin{tabular}{@{}c cc cc@{}}
\toprule
\textbf{Pretrain Mask Ratio (PMR)} & \multicolumn{2}{c}{\textbf{Inference at PMR}} & \multicolumn{2}{c}{\textbf{Inference at 75\% MR}} \\
\cmidrule(lr){2-3}\cmidrule(lr){4-5}
& \textbf{PSNR}~$\uparrow$ & \textbf{SSIM}~$\uparrow$ & \textbf{PSNR}~$\uparrow$ & \textbf{SSIM}~$\uparrow$ \\
\midrule
60\% & 38.84 & 0.9391 & 36.46 & 0.8961 \\
70\% & 38.12 & 0.9245 & 37.10 & 0.9015 \\
75\% & 37.55 & 0.9083 & \textbf{37.55} & \textbf{0.9083} \\
80\% & 36.52 & 0.8842 & 36.91 & 0.8937 \\
85\% & 34.21 & 0.8124 & 35.82 & 0.8684 \\
\bottomrule
\end{tabular}
\end{center}
\end{table}

\subsubsection{Masking Ratio Analysis}
\label{subsec:effect-mask-ratio}

Table~\ref{tab:masking-ratio-comparison} shows a 75\% pretrain masking ratio yields the best reconstruction for TerraMAE when evaluated at the same level. While lower ratios generalize poorly, higher ratios maintain strong performance under heavy masking.


\subsubsection{Impact of Spectral Grouping}
\label{subsec:impact-channel-groups}

Increasing spectral groups improves reconstruction but significantly raises training time (Table~\ref{table:ablation-study}). This reflects a trade-off between spectral granularity and computational cost.

\begin{table}[H]
\caption{California Test Set 2 performance with varying channel groups, evaluated at 75\% masking.}
\label{table:ablation-study}
\begin{center}
\scriptsize
\begin{tabular}{@{}c c c c c@{}}
\toprule
\textbf{Pretrain Mask Ratio} & \textbf{Groups} & \textbf{PSNR} $\uparrow$ & \textbf{SSIM} $\uparrow$ & \textbf{Time/Epoch} \\
\midrule
75\% & 5  & 37.55 & 0.9083 & \(\sim\)150s \\
75\% & 10 & 40.13 & 0.9267 & \(\sim\)280s \\
75\% & 20 & \textbf{42.47} & \textbf{0.9386} & \(\sim\)670s \\
\midrule
80\% & 5  & 36.91 & 0.8937 & \(\sim\)135s \\
80\% & 10 & 38.36 & 0.9081 & \(\sim\)250s \\
80\% & 20 & 40.02 & 0.9206 & \(\sim\)590s \\
\midrule
85\% & 5  & 35.82 & 0.8684 & \(\sim\)127s \\
85\% & 10 & 37.19 & 0.8816 & \(\sim\)230s \\
85\% & 20 & 38.22 & 0.8959 & \(\sim\)550s \\
\bottomrule
\end{tabular}
\end{center}
\end{table}

\end{document}